\documentclass[sigconf]{acmart}

\AtBeginDocument{%
  }

\copyrightyear{2025}
\acmYear{2025}
\setcopyright{acmlicensed}\acmConference[KDD '25]{Proceedings of the 31st ACM SIGKDD Conference on Knowledge Discovery and Data Mining V.2}{August 3--7, 2025}{Toronto, ON, Canada}
\acmBooktitle{Proceedings of the 31st ACM SIGKDD Conference on Knowledge Discovery and Data Mining V.2 (KDD '25), August 3--7, 2025, Toronto, ON, Canada}
\acmDOI{10.1145/3711896.3737256}
\acmISBN{979-8-4007-1454-2/2025/08}

\usepackage{amsfonts,amssymb}
\usepackage{color}
\definecolor{purplefish}{RGB}{138,43,226}
\definecolor{orangefish}{RGB}{210,105,30}
\definecolor{crimson}{RGB}{220,20,60}
\usepackage{mathrsfs}
\usepackage{url}
\usepackage{subfigure}
\usepackage{graphicx}

\usepackage{algorithmic}
\usepackage{algorithm}

\begin{document}

\title{Personalized Query Auto-Completion for Long and Short-Term Interests with Adaptive Detoxification Generation}


\author{Zhibo Wang}
\affiliation{
  \institution{Kuaishou Technology}
  \city{Beijing}
  \country{China}
}
\email{wangzhibo07@kuaishou.com}

\author{Xiaoze Jiang}
\authornote{Corresponding author.}
\affiliation{%
  \institution{Kuaishou Technology}
  \city{Beijing}
  \country{China}
}
\email{jiangxiaoze@kuaishou.com}

\author{Zhiheng Qin}
\affiliation{%
  \institution{Independent}
  \city{Beijing}
  \country{China}
}
\email{qinzhiheng1991@gmail.com}

\author{Enyun Yu}
\affiliation{%
  \institution{Independent}
  \city{Beijing}
  \country{China}
}
\email{yuenyun@126.com}

\author{Han Li}
\affiliation{%
  \institution{Kuaishou Technology}
  \city{Beijing}
  \country{China}
}
\email{lihan08@kuaishou.com}

\begin{abstract}
Query auto-completion (QAC) plays a crucial role in modern search systems. However, in real-world applications, there are two pressing challenges that still need to be addressed. First, there is a need for hierarchical personalized representations for users. Previous approaches have typically used users' search behavior as a single, overall representation, which proves inadequate in more nuanced generative scenarios. Additionally, query prefixes are typically short and may contain typos or sensitive information, increasing the likelihood of generating toxic content compared to traditional text generation tasks. Such toxic content can degrade user experience and lead to public relations issues. Therefore, the second critical challenge is detoxifying QAC systems. Recent efforts to mitigate toxicity have involved generating queries unrelated to the given prefix, leading this approach still negatively impacts user experience. 

To address these two limitations, we propose a novel model (LaD) that captures personalized information from both \textbf{l}ong-term and short-term interests, incorporating \textbf{a}daptive \textbf{d}etoxification. In LaD, personalized information is captured hierarchically at both coarse-grained and fine-grained levels. This approach preserves as much personalized information as possible while enabling online generation within time constraints. To move a futher step, we propose an online training method based on Reject Preference Optimization (RPO). By incorporating a special token {\ttfamily [Reject]} during both the training and inference processes, the model achieves adaptive detoxification. Consequently, the generated text presented to users is both non-toxic and relevant to the given prefix. We conduct comprehensive experiments on industrial-scale datasets and perform online A/B tests, delivering the largest single-experiment metric improvement in nearly two years of our product. Our model has been deployed on Kuaishou search, driving the primary traffic for hundreds of millions of active users. The code is available at \url{https://github.com/JXZe/LaD}.

\end{abstract}




\maketitle

\section{Introduction}

Query auto-completion (QAC) is a crucial feature in modern search systems. Generally speaking, given the query prefix (i.e. the substring of a word or a sentence), the search engines can recommend several candidate queries to users. The recent development of the generative pre-trained language model \cite{baek24kllqs,zhong2020personalized} has inspired a new surge of interest in QAC generation \cite{maheswaran2024dac,gupta2023deep,yin2020learning,li2025uctg}. These works focus on designing complex methods to improve the QAC generative ability on public datasets or to conduct near-line experiments in industrial applications \cite{yin2020learning}. Meanwhile, in real-world scenarios, QAC faces even more formidable challenges according to users' satisfaction. On one hand, the prefix tends to be incomplete, often contain spelling errors etc. But the provided queries should literally be related to prefix, contain fewer grammatical and spelling errors, and avoid politically or sexually sensitive content (which can be defined as detoxification \cite{maheswaran2024dac,maheswaran2024dqac}). On the other hand, the QAC engine should satisfy the personalized search intents of various users when the prefixes are the same. Therefore, equipping QAC with personalized capturing and detoxification capabilities for online generation has become a critical yet underexplored challenge for real-world applications.

\begin{figure} [t]
\centering
\includegraphics[width=8.3cm]{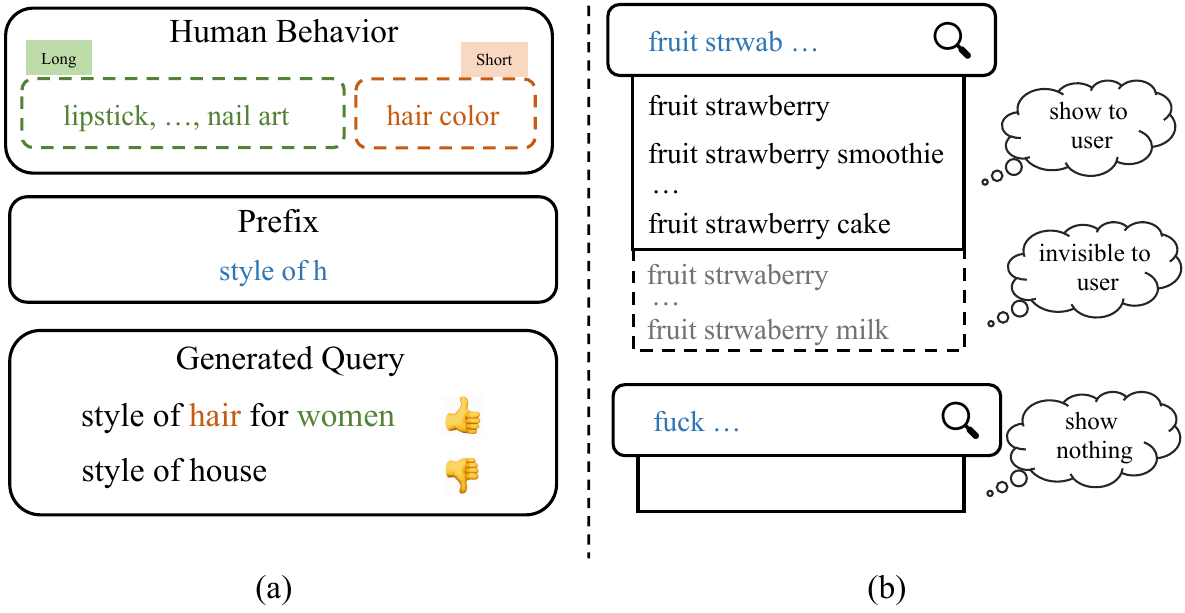}
\caption{A schematic illustration of our method LaD. (a) Given the user's behavior and the prefix, our model can generate completions from the long- (\emph{women}) and short-term (\emph{hair}) interests respectively; (b) Under the toxic prefix (\emph{fruit strwab} and \emph{fuck}), our model can reply with non-toxic contents adaptively.}
\label{fig:intro}
\end{figure}

To characterize personalized features, some previous studies \cite{zhong2020personalized,bao2024search,yin2020learning} tried to encode the user's behavior into a vector and then feed it into a decoder. Some other works \cite{maurya2023trie,baek24kllqs} introduced additional knowledge from common web-pages or candidate queries. However, they overlooked the need to balance the modeling of user's long-term and short-term interests. For example, as shown in Figure \ref{fig:intro}.a, before typing the prefix `` \emph{style of h}'' , the nearest pre-searched query is `` \emph{Hair Color}''. It captures the user's short-term interests, which should be interacted with the prefix in a fine-grained manner.  `` \emph{Lipstick, …, Nail Art}'' are frequently searched by the user or are spaced further apart in time from the prefix. It reflects the user's long-term interests, which can be recognized coarsely to store the user's basic information. Then the awesome query `` \emph{style of hair for women}'' can be generated (`` \emph{hair}'' is from the short-term interests, while `` \emph{women}'' is from the long-term).  A straightforward approach is to incorporate all human behaviors (such as search queries) into the model. Nevertheless, this may involve two main drawbacks. The first is that the human behavior terms may be too lengthy for the model to encode, making it unrealistic for online generation. Secondly, extended term features may introduce more noise, potentially hindering fine-grained generation.

After we effectively capture user interests, another issue arises. Due to the uncontrollability of generative models, they tend to produce toxic text, especially when the prefix is already toxic. This may result in challenges for the online user experience, rendering generative methods incapable of performing real-time inference. As for the solution to detoxifying QAC, traditional methods involve user feedback, filter toxic contents by the rule or conduct generation upon limited query templates \cite{hazen2020social,gibbs2016google,davidson2017automated,silva2016analyzing}. These efforts are unsuccessful in achieving flexible generation and require constant maintenance. There are also some works that focus on non-toxic text generation for long texts by detoxifying clean datasets \cite{gururangan2020don}, refining the decoding process \cite{krause2021gedi,dathathri2020plug}, or introducing reinforcement learning based approaches \cite{lu2022quark,wu2023fine}. Since the prefixes and completions are often short, misspelled or imcomplete in QAC system, rendering these models ineffective for QAC detoxification tasks. Recent study \cite{maheswaran2024dac,maheswaran2024dqac} can generate non-toxic queries without considering the relevance between the prefix and the completions (i.e. the model tends to generate irrelevant completions when the prefix is toxic). As shown in \ref{fig:intro}.b, in real-world applications, when the prefix is `` \emph{fruit strwab}'' (containing spelling errors), the ideal completions (such as `` \emph{fruit strawberry}'') should be displayed to users, while any toxic text (such as `` \emph{fruit strwaberry milk}'') should remain invisible to them. When the prefix is `` \emph{fuck}'' (containing sexual innuendo), the QAC system should return no results. In other words, while maintaining relevance, prefixes containing politically or sexually sensitive information should not display any output. Thus, implementing adaptive detoxification generation for various types of prefixes is crucial to practical QAC systems, yet it has been rarely explored in previous research.


To address the aforementioned limitations, we propose LaD, a model designed to capture both \textbf{L}ong- and short-term interests, along with \textbf{a}daptive \textbf{D}etoxification in query auto-completion generation (QAC). LaD depicts user's interests in a hierarchical structure. The first level is coarse-grained, storing long-term interests, while the second level is fine-grained, reflecting short-term interests. Specifically, long-term interests are encoded with several sentence-level representations based on the user's frequent or past behaviors. Short-term interests are modeled by the adjacent search behaviors with token-level embeddings. These representations are then input into the encoder along with the prefix to enable interaction among them. To achieve adaptive detoxification in generation, LaD is implemented using an end-to-end method. In detail, LaD employs a new special token, {\ttfamily [Reject]}, during the decoding process. In training stage, the {\ttfamily[Reject]} is inserted into the generated sequence by a Detoxification Expert (is implemented as a discriminative model), which can evaluate the quality of the text. During inference, any sequence sorted lower than the {\ttfamily[Reject]} token is discarded from the final results. Meanwhile, the model's optimization strategy, called Reject
Preference Optimization (RPO), is meticulously crafted to ensure that the generated queries ranked ahead of the {\ttfamily[Reject]} token are entirely non-toxic. As a result, the model is able to adaptively complete query based on different inputs.

There are four main contributions in our work: 

\begin{figure*}[t]
\centering
\includegraphics[width=17cm]{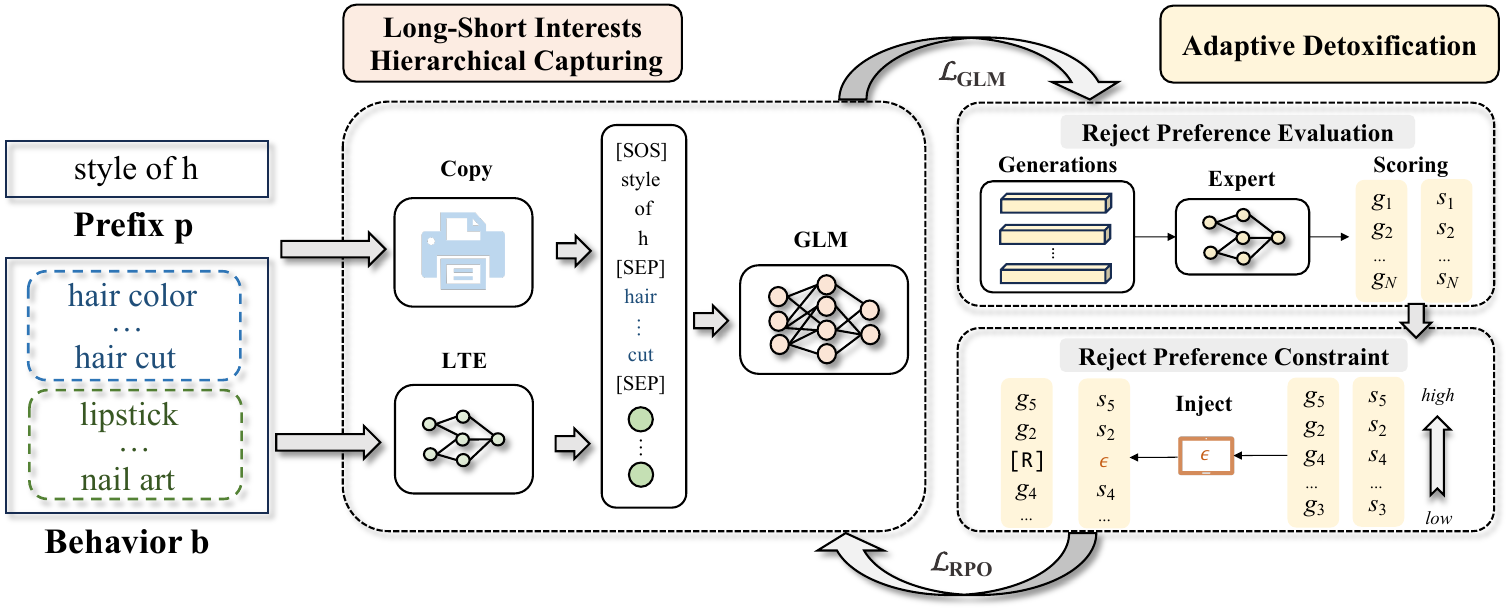}
\caption{Overall structure of the LaD, where ``{\ttfamily [R]}'' denotes the special token {\ttfamily [Reject]}, ``LTE''  denotes Long-term interests Transformer Encoder and ``GLM''  denotes Generative Language Model. The model primarily consists of two components: \emph{Long-Short Interests Hierarchical Capturing} and \emph{Adaptive Detoxification}.}
\label{fig:model_pic}
\end{figure*}

(1) We emphasize the two limitations of QAC generation task for online applications. The first is the lack of hierarchical interest representation, and the second is the inability to adaptively generate non-toxic completions. The advanced personalized QAC industry generation method should incorporate detoxification capabilities, as toxic text can degrade the user experience, cause public relations challenges, and render the system unfit for deployment. 

(2) As the first attempt, we achieve a hierarchical interest characterization framework, called LaD, equipped with adaptive detoxification generation capability. It captures both long-term and short-term user interests, as well as adaptively generating non-toxic text based on the different types of prefix input. 

(3) Distinct from other works, our model can deliver authentic online real-time inference while adhering to time constraints. It can transcend the limitations of indexing, offering faster and more effective support for users' search intentions than the near-line or offline methods. 

(4) We validate LaD on Kuaishou search. Offline experimental results demonstrate that LaD excels in personalized characterization and adaptive detoxification generation. In online A/B tests, our method outperforms the strong production baseline and obtains substantial enhancement on \emph{Click-through Rate} and \emph{Search Penetration Rate}. LaD has now been launched on Kuaishou search, serving the primary traffic for hundreds of millions of active users.


\section{RELATED WORK}


\noindent \textbf{Query Auto-Completion (QAC).} 
QAC has become an indispensable feature of search engines for improving user experience and reducing user input costs. Early methods for QAC relied on association rules \cite{fonseca2003using} and co-occurrence information \cite{fonseca2005concept,huang2003relevant} such as Most Popular Completion (MPC) \cite{bar2011context} which suggests the most popular queries starting with the given prefix. In contrast to previous methods relying on a query candidate pool, sequence-to-sequence generation methods are applied in query auto-completion \cite{fiorini2018personalized,wang2018realtime}, leading to better handling of unseen prefixes. Inspired by the success of the attention mechanism \cite{vaswani2017attention,jiang2022xlm} in various NLP tasks,
transformer-based QAC models \cite{mustar2020using,yin2020learning} have been shown to be more robust to noise and can generate more diverse results. To improve the grasp of user search intent, personalized information is brought into play such as search session information \cite{maurya2023trie} and user behavior \cite{bao2024search}. In our work, we break down user historical behavior into long- and short-term interests and create a hierarchical model that maximizes the utilization of user historical behavior information while meeting the requirements for online inference efficiency.

\noindent \textbf{Detoxification in QAC.} 
Some studies have focused on detoxification in free-form generation by detoxifying clean datasets \cite{maity2024multilingual,gururangan2020don}, refining the decoding process \cite{krause2021gedi,dathathri2020plug}, or employing reinforcement learning-based approaches \cite{lu2022quark,wu2023fine}. However, since prefixes and completions in QAC systems are often short, misspelled, or incomplete, these models are ineffective for QAC detoxification tasks. Detoxification methods in QAC can be broadly classified into three categories: rule-based methods \cite{hazen2020social,gibbs2016google,davidson2017automated,silva2016analyzing}, learning-based approaches \cite{huang2013learning,settles2009active,chuklin2013adult,yenala2017convolutional,gibbs2016google}, and reinforcement learning-based techniques \cite{maheswaran2024dqac,maheswaran2024dac}. The rule-based methods demand substantial and continuous human effort over a prolonged period. The learning-based approaches employ a two-stage detoxification process, where the retrieved queries are first evaluated for toxicity and then filtered accordingly. The reinforcement learning-based techniques, in an effort to generate non-toxic text, often tend to produce completions that are unrelated to the prefix. Our model achieves end-to-end adaptive detoxification, ensuring that all generated outputs remain relevant to the given prefix. Furthermore, when encountering sensitive information, our QAC model returns no output. With its robust adaptive detoxification capabilities, the model supports real-time online generation.

\section{METHODOLOGY}

The personalized QAC can be described as follows \cite{yin2020learning,maheswaran2024dac}: given a user $u$ entering the prefix $p$ and the user's search behaviors (containing past searched queries $B_{l} = \{ b_{l1}, b_{l2}, ..., b_{lL}\}$ and the recent searched queries $B_{s}= \{ b_{s1}, b_{s2}, ..., b_{sS}\}$). The task is to generate $N$ candidate completions $G = \{ g_{1}, g_{2}, ..., g_{N}\}$. In this section, we will present our LaD model, featuring the concepts of {\it Long-Short Interests Hierarchical Capturing} and {\it Adaptive Detoxification}. Following this, we will discuss the online generation scheme of LaD.


\subsection{Long-Short Interests Hierarchical Capturing}
\label{sec:lshc}

The user's behavior is crucial for responding to their prefix in a search engine. Traditional methods \cite{zhong2020personalized,bao2024search,yin2020learning} address this factor in a coarse-grained manner by encoding all behaviors into one or several vectors, which are then used for decoding. It overlooks the term-level features stored in the user's previous search queries, which are crucial for fine-grained personalized generation. In other fields, capturing approaches vary: some \cite{ye2020lsan} model the same sequence twice, while others \cite{xu2021long} incorporate short-term pattern capture within the decoder. These methods fail in online QAC due to latency restrictions. Intuitively, incorporating all the words from the behaviors into the model can capture the maximum amount of information. However, it is impractical for real-world online generation scenarios with limited server resources and urgent real-time response speed. To move a further step, we capture user's interests in a hierarchical structure.

\noindent \textbf{Long-Term Interests Encoding} 

\noindent It captures the user's long-term interests, including gender, preferred genres of films and TV shows, frequently used phrases, and more. Since the long-term interests remain unchanged over an extended period despite different prefix, they can be encoded in a coarse-grained manner. As shown in Figure \ref{fig:model_pic}, we assign a \textbf{T}ransformer \textbf{E}ncoder \cite{vaswani2017attention} to store the user's \textbf{L}ong-term interests (denotes LTE, in Figure \ref{fig:model_pic}):
\begin{align}
\widetilde{b}_{li} = \textnormal{LTE}([{b}_{li1},{b}_{li2},...,{b}_{lin}])
\end{align}
where $b_{lij}$ is the $j$-th ($j=1,...,n$) term from $i$-th user's past searched query $b_{li}$ from the $B_{l}$, and $\widetilde{b}_{li}$ is the encoded long-term representation of  $b_{li}$ ($i=1,...,L$, $L$ is the length of $B_{l}$).

\noindent \textbf{Short-Term Interests Encoding}

\noindent Unlike long-term interests, short-term interests reflect the user's immediate feedback. They precisely capture the user's immediate search intentions, such as finding a specific episode of a TV show or movie series, seeking restaurant reviews for a particular location, or following a specific breaking news event, among other interests. To address these refined needs, sophisticated model designs are required to capture and predict them accurately. As shown in Figure \ref{fig:model_pic}, we develop a copying method to represent short-term interests:
\begin{align}
[{b}_{si1},{b}_{si2},...,{b}_{sin}] = \textnormal{Copy}([{b}_{si1},{b}_{si2},...,{b}_{sin}])
\end{align}
where $b_{sij}$ is the $j$-th ($j=1,...,n$) term from $i$-th user's recent searched query $b_{si}$ from the $B_{s}$ ($i=1,...,S$, $S$ is the length of $B_{s}$).

\noindent \textbf{Personalized Generation}

\noindent After establishing a hierarchical representation of user interests, another challenge arises: how to integrate long-term and short-term interests. Fortunately, the extensive research on multi-head attention (MHA) provides a natural framework for encoding different information based on varying inputs. Thus, we feed the prefix $p$, long-term interests $\widetilde{b}_{li}$ and short-term interests $b_{sij}$ into a \textbf{G}enerative \textbf{L}anguage \textbf{M}odel (denotes GLM):
\begin{align}
\mathbb{P}=[p_{1},...,{p}_{p}]&\\
\mathbb{L}=[\widetilde{b}_{l1},...,\widetilde{b}_{lL}]&\\
\mathbb{S}=[{b}_{s11}, ...,{b}_{s1n},b_{s21},...,{b}_{sSn}]&\\
[g_1,...,g_N] =\textnormal{GLM}([\mathbb{P},\mathbb{S},\mathbb{L}])&
\end{align}
where $p_{p}$ is the $p$-th terms of $p$, $\widetilde{b}_{lL}$ is the embeddings from the $L$-th of long-term interests  $\widetilde{b}_{l}$, ${b}_{sSn}$ is the $n$-th terms of the $S$-th of short-term interests $b_{s}$ and $g_{N}$ is the $N$-th generation from GLM. The GLM can be implemented in two architectures: an encoder-decoder \cite{lewis-etal-2020-bart} model and a decoder-only \cite{floridi2020gpt} model. We conduct our experiments using both types of architectures in Section \ref{Experiments} and select the better one for detailed analysis and ablation experiments. The aim of the optimization is to minimize the following objective:
\begin{align}
\mathcal{L}_{GLM} = -\textnormal{log}  \sum_{(y,[\mathbb{P},\mathbb{S},\mathbb{L]})} P(y|[\mathbb{P},\mathbb{S},\mathbb{L}])
\label{eq:loss_glm}
\end{align}
where $y$ is the ground truth.

\subsection{Adaptive Detoxification}
As generative technologies become more widely adopted, increasing attention is being paid to the quality issues of generated data. In QAC tasks, these challenges are particularly pronounced, as the prefixes and completions are often short, misspelled, or incomplete. Some attempts \cite{hazen2020social,davidson2017automated} have fallen short of meeting the requirements for flexible prefixes, while others \cite{maheswaran2024dac,maheswaran2024dqac} tend to produce unrelated, non-toxic text. Furthermore, the literal quality of completions is vital to the user experience. The generation of toxic text prevents generative QAC from being implemented and may even pose public relations risks. To address the issues mentioned above, we propose a method capable of achieving adaptive detoxification, called Reject Preference Optimization (RPO). Under the RPO training scheme, our model can deliver relevant and non-toxic completions to users. In other words, the model can adaptively generate content tailored to various prefixes. For instance, when the prefix contains a significant amount of sensitive information (such as content related to pornography and political sensitivity), the model produces no output. In contrast, if the prefix is only misspelled or incoherent, the model is capable of generating text that is both non-toxic and contextually relevant. The RPO has two parts: {\it Reject Preference Evaluation} and {\it Reject Preference Constraint}.

\noindent \textbf{Reject Preference Evaluation}

\noindent In QAC industry applications, the standard approach to handling toxic candidate queries involves leveraging a discriminative model or a grammatical error correction model \cite{katinskaia-yangarber-2023-grammatical,katinskaia2024gpt} for post-processing. Deploying an additional model after generation will incur extra latency for real-time online applications. 

As illustrated in Figure \ref{fig:model_pic}, to achieve end-to-end online detoxification generation, we first introduce a Detoxification Expert. This expert is implemented as a discriminative transformer model, allowing for the immediate evaluation of the quality of completions as they are generated during training. The Detoxification Expert is trained using a dataset comprising hundreds of millions of search log entries and tens of thousands of manually annotated samples. In each training step, the Detoxification Expert ranks the generated list $G=\{g_1,...,g_N\}$ in descending order, where $N$ represents the number of candidate queries generated for each prefix.



\noindent \textbf{Reject Preference Constraint}

\noindent The quality of generated completions plays a critical role in shaping user experience, while the handling of sensitive information is essential for maintaining platform security. Consequently, RPO is designed to suppress toxic text while ensuring relevance, rather than focusing on exploring more refined expressions, such as RHLF-related methods \cite{ouyang2022training,ziegler2019fine}.

In detail, we set a threshold $\epsilon$ in collaboration with Detoxification Expert. The special token {\ttfamily [Reject]}
is injected into the sorted generations $G$  when they fall below the threshold $\epsilon$ (shown in Figure \ref{fig:model_pic}). The Reject Preference Constraint can be calculated as:
\begin{multline}
\mathcal{L}_{RPO} = -(\sum_{(y_+,y_r)} \textnormal{log} \ \sigma \ [\textnormal{log} \  P(y_{+}|[\mathbb{P},\mathbb{S},\mathbb{L}]) - \textnormal{log} \ P(y_{r}|[\mathbb{P},\mathbb{S},\mathbb{L}])] + \\
\sum_{(y_r,y_-)} \textnormal{log} \ \sigma \ [\textnormal{log} \  P(y_{r}|[\mathbb{P},\mathbb{S},\mathbb{L}]) - \textnormal{log} \ P(y_{-}|[\mathbb{P},\mathbb{S},\mathbb{L}])])
\label{eq:loss_rpo}
\end{multline}
where $y_r$ is the notation of {\ttfamily [Reject]} and $y_{-}$ ($y_{+}$) is the generations discarded (boosted) by the Detoxification Expert. Furthermore, the RPO requires that the probability of generating  {\ttfamily [Reject]} is absolutely better than that of the subsequent candidates, in order to achieve adaptive generation. This is different from DPO-related algorithms \cite{rafailov2024direct,khaki2024rs}, which require the generated result to be relatively better than the reference model.

\begin{algorithm}[t]
    \caption{Algorithm of the LaD}
    \label{alg:AWM}
    \renewcommand{\algorithmicrequire}{\textbf{Input:}}

    \begin{algorithmic}[1]
        \REQUIRE prefix $p$, long-term interests $b_{li}$, short-term interests $b_{si}$,  Detoxification Expert $\mathbb{E}$, detoxification threshold $\epsilon$ and training step $T$. 
        
        
        
        
        \FOR{each step $t \in [1,T]$}
            \FOR{mini step $k \in [1,2]$}
                \STATE $\mathbb{P} \gets [p_1,...,p_p]$
                \STATE $\mathbb{L} \gets \widetilde{b}_{li} \gets \textnormal{LTE}([{b}_{l1},...,{b}_{li}])$
                \STATE $\mathbb{S} \gets [{b}_{s11},...,{b}_{sSn}] \gets \textnormal{Copy}([{b}_{s11},...,{b}_{sSn}])$
                           
                \IF {$k = 1$}
                    \STATE $[g_1,...,g_N] \gets \textnormal{GLM}([\mathbb{P},\mathbb{S},\mathbb{L}])$ 
                    \STATE $[s_1,...,s_N] \gets \mathbb{E} \gets [g_1,...,g_N]$
                    \IF {$ s_{a} < \epsilon < s_{b}$}
                        \STATE $[...,g_{b}, ${\ttfamily [Reject]} $,g_{a} ...] \gets$ inject {\ttfamily [Reject]}
                    \ENDIF
                \ELSE
                    \STATE   $\mathcal{L}_{GLM}, \ \mathcal{L}_{RPO} \gets \textnormal{GLM}([\mathbb{P},\mathbb{S},\mathbb{L}])$ using Eq. (\ref{eq:loss_glm}),  (\ref{eq:loss_rpo})
                \ENDIF
            
            \ENDFOR            
            \STATE Backpropagation: $\nabla \textnormal{GLM} + \nabla \textnormal{LTE} \gets \mathcal{L}_{GLM} + \ \mathcal{L}_{RPO}$  
        \ENDFOR
        
    \end{algorithmic}
\end{algorithm}

\noindent \textbf{Overall Training Scheme}

\noindent The whole model is trained with generation loss $\mathcal{L}_{GLM}$ and RPO constraint  $\mathcal{L}_{RPO}$:
\begin{align}
\mathcal{L} = \mathcal{L}_{GLM} + \mathcal{L}_{RPO}
\end{align}
where $\mathcal{L}_{GLM}$ optimizes the model to improve its capacity for personalized generation, while $\mathcal{L}_{RPO}$ equips the model with adaptive detoxification capabilities. The algorithmic process is illustrated in Algorithm \ref{alg:AWM}. It is worth noting that the {\ttfamily [Reject]} token is integrated into the {\it  \textbf{online training process}} instead of being applied during the preprocessing of the data's ground truth. The candidate queries from the datasets are not truly generated by the model. Therefore, whether their generation probabilities are higher or lower than that of {\ttfamily [Reject]}, it will not significantly affect the generated results. The real-time insertion of {\ttfamily [Reject]} at each training step allows for faster and more precise capture of the model's output distribution, thus achieving genuine insertion.

\subsection{Online Generation}
\label{sec:online_gene}

Currently, numerous studies in the QAC field are conducted using offline datasets \cite{maheswaran2024dqac} or through nearline experiments \cite{yin2020learning}. In contrast to these approaches, our model is capable of providing genuine online real-time inference while meeting time constraints.

\begin{figure}
\centering
\includegraphics[width=6cm]{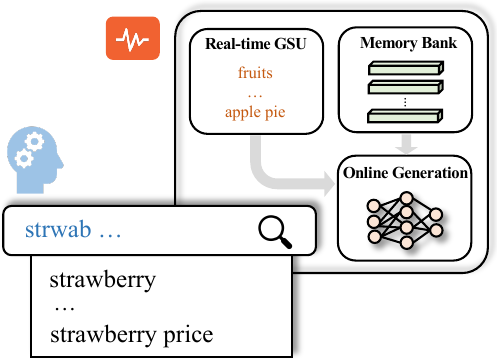}
\caption{An illustration of online generation.}
\label{fig:online}
\end{figure}

As illustrated in Figure \ref{fig:online}, the inputs for online serving consist of three components: the user's prefix, the Real-time GSU (which provides the user's recently searched queries), and the Memory Bank (which stores the user's long-term representations and is daily updated using LTE). The model uses these inputs to generate non-toxic completions for the user. In practice, any generated outputs with a score lower than that of the special token {\ttfamily [Reject]} will not be shown to the user. The generated results currently act as a new source for recall, and could be presented as the final output to users in the future. Online generation helps us overcome the limitations of indexing. It captures users' search intent in real time, enhances the user experience, and cultivates a search mindset among users.

\section{Experiments}
\label{Experiments}

\subsection{Implementation Details}
\noindent \textbf{Dataset}

\noindent The existing public datasets do not address the practical engineering challenges we face.  Some datasets lack real prefix-to-query click behaviors \cite{pass2006picture,everaert2024amazonqac}, while others contain only limited personalized information \cite{yin2020learning}. In real-world industrial development, user behavior information is extremely rich and diverse, characterized by long sequences. Additionally, a substantial amount of prefix-to-query click behavior data is generated daily. Therefore, there is a clear gap between most previous research work and real-world industrial deployment. 

To more effectively tackle the engineering challenges we encounter, we develop a QAC dataset specifically tailored to the domain of Kuaishou search. We name this dataset as KSQAC. The KSQAC dataset is collected from online query logs spanning December 1, 2024, to December 28, 2024. The detailed statistics can be found in  Table \ref{tabel:data}.




\noindent \textbf{Details of the LaD}

\begin{table}[t] 
\centering
\caption{The statistics of the KSQAC dataset. ``K'', ``M'', ``B'' indicate thousand, million, and billion respectively.}
\label{tabel:data} 
\resizebox{0.8\columnwidth}{!}{
\begin{tabular}{lcc} 
\hline                       
Split Type & Training & Testing  \\
\hline  
{Number of queries} & {40M} & {4K}  \\
{Number of users} & {11M} & {1K}  \\
{Number of samples} & {100M} & {10K}  \\
{Number of search behaviors} &{2.9B} & {289K}   \\
\hline  
\end{tabular}  
}
\end{table} 

\noindent Our model, LaD, as well as all the models we compared it against, are implemented using PyTorch.  We utilize Adam \cite{kingma2015adam} as our optimizer with the gradient accumulation of 512 batch size. The learning rate starts with 10k warm-up steps and the peak learning rate is set to 3e-5. We adopt an encoder-decoder architecture \cite{mustar2020using} as the core implementation. Both the encoder and decoder have a hidden size of 768, use 12 attention heads, and consist of 6 layers. The length of long-term interests $L$ is 7, and the length of long-term interests $S$ is 3. LTE is implemented as an 8-layer Transformer encoder. The Detoxification Expert is implemented as a 48-layer Transformer encoder. Then, it is trained with point-wise constraints, ensuring its output scores have inherent physical meaning, where higher scores directly indicate better textual quality. The rejection threshold, denoted as $\epsilon$ and illustrated in Figure \ref{fig:model_pic}, is set to 0.6, which is precisely calibrated to maintain both the precision and recall rates of our Detoxification Expert above 90\textsl{\%}. For each prefix, LaD generates up to 4 completions. For the KSQAC dataset, the vocabulary is constructed using Chinese characters, resulting in a total vocabulary size of 53,704. The experiments are conducted using 8 V100 GPUs. The latency of our online LaD is 20 ms (much lower than the 30 ms time limit).


\noindent \textbf{Evaluation Metrics}

\noindent {\it \textbf{Generation Metrics.}}  We take into account the model's recall, ranking and generation performance. To evaluate recall performance, we use Recall@4 (R@4) as the metric. Let $Num_{p+}$ as the number of instances predicted as positive and labeled as positive among the top-4 retrieval results, $Num_+$ as the number of instances labeled as positive, $N$ represents the number of samples:
\begin{align}
\textnormal{R@4} = \frac{1}{N} \sum_{i=1}^{N} \frac{Num_{p+}}{Num_+}
\end{align}
To evaluate ranking performance, we utilize MRR as the metric. Let $R_i$ be a positive number that indicates the position of the golden query for the $i$-th sample among the ranked candidates:
\begin{align}
\textnormal{MRR} = \frac{1}{N} \sum_{i=1}^{N} \frac{1}{R_i}
\end{align}

\noindent To evaluate generation performance, we use BLEU as the metric. The details of BLEU can be found in \citeauthor{papineni2002bleu}'s work \cite{papineni2002bleu}.

\noindent {\it \textbf{Detoxification Metrics.} }To assess the performance of detoxification, we introduce the  Average Max Toxicity (AmaxT, the average of the maximum toxicity over 4 generations for a test example) \cite{gehman2020realtoxicityprompts} and the Empirical Toxicity Probability (Prob, the probability that at least one of the 4 generations is toxic) \cite{gehman2020realtoxicityprompts}. In the context of adaptive generation, this metric is biased because as the number of examples generated decreases, there are fewer toxic queries, which can result in better metric scores. To penalize the generation of fewer queries, we designed the Unbiased AmaxT and Prob (denotes UAmaxT and UProb):

\begin{align}
\textnormal{UAmaxT} = \frac{N_G}{\frac{1}{N} \sum_{i=1}^{N} N_{gi}} \textnormal{AmaxT}, \, \ 
\textnormal{UProb} = \frac{N_G}{\frac{1}{N} \sum_{i=1}^{N}  N_{gi}} \textnormal{Prob} 
\end{align}
where $N_{gi}$ is the length of generations after detoxification, while $N_G$ ($=4$ here) is the maximum generation length. Since our model can adaptively filter out toxic generations, we report the average number of rejected generations, AvgRN, for ablation study.

\subsection{Overall Results}

We first conduct experiments on the KSQAC dataset (as mentioned above). The results are described as follows.

\noindent \textbf{Comparative Methods}

\begin{table}[t] 
\centering
\caption{The Results of LaD on KSQAC Test dataset, where ``AD'' denotes adaptive detoxification and ``De'' denotes decoder only model.}
\label{tabel:exp_all} 
\resizebox{0.95\columnwidth}{!}{
\begin{tabular}{l|ccc|cc} 
\hline 
  & R\textsl{@}4 $\uparrow$
 & BLEU $\uparrow$
 & MRR $\uparrow$
 &  UAmaxT $\downarrow$
 & UProb $\downarrow$
 \\
\hline  
\hline 
MPC \cite{bar2011context} & 22.32\textsl{\%} & 18.54\textsl{\%} & 15.77\textsl{\%} & 0.2109 & 26.31\textsl{\%}  \\
Gen \cite{mustar2020using} & 26.11\textsl{\%} & 31.95\textsl{\%} & 20.27\textsl{\%} & 0.1746 & 22.77\textsl{\%}  \\
M$^{2}$A \cite{yin2020learning} & 27.87\textsl{\%} & 32.60\textsl{\%} & 21.54\textsl{\%} & 0.1782 & 23.11\textsl{\%}  \\
SIN \cite{bao2024search} & 28.04\textsl{\%} & 32.65\textsl{\%} & 21.56\textsl{\%} & 0.1782 & 23.23\textsl{\%}  \\
Trie-NLG \cite{maurya2023trie} & 30.60\textsl{\%} & 34.55\textsl{\%} & 23.52\textsl{\%} & 0.1802 & 23.23\textsl{\%}  \\

\hline  
\hline
DAPT \cite{gururangan2020don} & 31.48\textsl{\%} & 34.80\textsl{\%} & 24.71\textsl{\%} & 0.1219 & 15.95\textsl{\%}  \\
Quark \cite{lu2022quark} & 22.11\textsl{\%} & 26.34\textsl{\%} & 15.56\textsl{\%} & 0.0948 & 12.47\textsl{\%}  \\
PPO \cite{ziegler2019fine} & 30.65\textsl{\%} & 33.13\textsl{\%} & 23.80\textsl{\%} & 0.1102 & 14.16\textsl{\%}  \\
DAC \cite{maheswaran2024dac} & 29.69\textsl{\%} & 32.19\textsl{\%} & 22.99\textsl{\%} & 0.1043 & 13.19\textsl{\%}  \\
\hline  
\hline
LaD w/o AD & \textbf{31.70\textsl{\%}} & \textbf{35.15\textsl{\%}} & \textbf{24.78\textsl{\%}} & 0.1400 &  19.31\textsl{\%}   \\ 
LaD & 29.42\textsl{\%} & 31.77\textsl{\%}  & 23.09\textsl{\%} & \textbf{0.0590} & \textbf{6.55\textsl{\%}}  \\
\hline
\hline
LaD-De w/o AD & 31.62\textsl{\%} & 33.96\textsl{\%} & 24.56\textsl{\%} & {0.1579} & {20.63\textsl{\%}}  \\
LaD-De & 30.83\textsl{\%} & 33.34\textsl{\%} & 23.94\textsl{\%} & {0.0727} & {9.17\textsl{\%}}  \\
\hline
\end{tabular}  
}
\end{table} 

\noindent The compared models are shown in Table \ref{tabel:exp_all}, which can be divided into two major categories. (1) The models in the first block of Table \ref{tabel:exp_all} are the models without considering of detoxification. In detail, we compare our model with statistical methods such as \textbf{MPC} \cite{bar2011context}, generative models that do not incorporate personalized capturing, like \textbf{Gen} \cite{mustar2020using}, and generative models that introduce personalized capturing but lack hierarchical capturing, such as \textbf{M$^{2}$A} \cite{yin2020learning}, \textbf{SIN} \cite{bao2024search} and \textbf{Trie-NLG} \cite{maurya2023trie}. 
(2) The models listed in the second part of Table \ref{tabel:exp_all} are equipped with detoxification capabilities. We compare our approach with methods that are trained using non-toxic queries, such as \textbf{DAPT} \cite{gururangan2020don}, and those utilizing reinforcement learning algorithms, including \textbf{Quark} \cite{lu2022quark}, \textbf{PPO} \cite{ziegler2019fine}, and \textbf{DAC} \cite{maheswaran2024dac}.

\noindent \textbf{Analysis of the Overall Results}

\noindent As shown in Table \ref{tabel:exp_all}, the key analysis can be concluded as follows: 
(1) \textbf {The trade-off between toxicity reduction and performance}. Models with detoxifying capabilities (such as \textbf{DAC} and \textbf{LaD w/o AD}) tend to perform lower in metrics like R\textsl{@}4, BLEU, and MRR, but achieve better results in UAmaxT and UProb. This trade-off is especially pronounced in the QAC task due to the typically short nature of completions, which often consist of only a few words or even parts of words. 
(2) \textbf {The scalability of the structure}. As shown in Figure \ref{fig:model_pic} and discussed in Section \ref{sec:lshc}, \textbf{LaD} can be equipped with different types of generative language models. We implemented two variations: the first is an encoder-decoder model, \textbf{LaD}, and the second is a decoder-only model, \textbf{LaD-De}. These two models exhibited similar performance, but \textbf{LaD} performed slightly better on detoxification metrics. We choose the better-performing model, \textbf{LaD}, for subsequent analysis and online experiments. 
(3) \textbf{The advancement of the model LaD}. \textbf{LaD} achieves comparable results on generation metrics, and delivers the best performance on detoxification metrics among detoxified models (as shown in the second part of Table \ref{tabel:exp_all}), particularly outperforming reinforcement learning-based methods such as \textbf{PPO} \cite{ziegler2019fine}. Furthermore, our model without adaptive detoxification, \textbf{LaD w/o AD}, surpasses other models in the personalized QAC category (as shown in the first part of Table \ref{tabel:exp_all}). These results demonstrate that our model not only excels at capturing user intent but also achieves adaptive non-toxic generation, providing a safeguard for secure online applications. 
What's more, the results of LaD on public dataset can be found in Appendix.

\begin{table}[t] 
\centering
\caption{Performance comparison for Toxic Test Set ($T_{test}$) of KSQAC dataset, where ``AD'' denotes adaptive detoxification.}
\label{tabel:KSQAC_toxic} 
\resizebox{0.95\columnwidth}{!}{
\begin{tabular}{l|ccc|cc} 
  
\hline 
  & R\textsl{@}4 $\uparrow$ & BLEU $\uparrow$ & MRR $\uparrow$ &  UAmaxT $\downarrow$ & UProb $\downarrow$ \\
\hline  
\hline 
MPC \cite{bar2011context} & 16.62\textsl{\%} & 16.37\textsl{\%} & 11.89\textsl{\%} & 0.4594 & 60.34\textsl{\%}  \\
Gen \cite{mustar2020using} & 10.53\textsl{\%} & 23.04\textsl{\%} & 7.38\textsl{\%} & 0.3574 & 48.65\textsl{\%}  \\
M$^{2}$A \cite{yin2020learning} & 10.18\textsl{\%} & 23.70\textsl{\%} & 7.14\textsl{\%} & 0.3741 & 49.52\textsl{\%}  \\
SIN \cite{bao2024search}  & 9.83\textsl{\%} & 23.94\textsl{\%} & 7.04\textsl{\%} & 0.3715 & 49.43\textsl{\%}  \\
Trie-NLG \cite{maurya2023trie}  & 14.19\textsl{\%} & 26.46\textsl{\%} & 10.56\textsl{\%} & 0.3911 & 52.13\textsl{\%}  \\

\hline  
\hline
DAPT \cite{gururangan2020don}  & 11.40\textsl{\%} & 23.62\textsl{\%} & 7.68\textsl{\%} & 0.2835 & 39.86\textsl{\%}  \\
Quark \cite{lu2022quark}  & 6.01\textsl{\%} & 17.25\textsl{\%} & 4.12\textsl{\%} & 0.2336 & 31.77\textsl{\%}  \\
PPO \cite{ziegler2019fine} & 9.49\textsl{\%} & 21.70\textsl{\%} & 6.04\textsl{\%} & 0.2650 & 37.25\textsl{\%}  \\
DAC \cite{maheswaran2024dac}  & 7.14\textsl{\%} & 19.10\textsl{\%} & 4.65\textsl{\%} & 0.2294 & 33.25\textsl{\%}  \\
\hline  
\hline
LaD w/o AD & 13.58\textsl{\%} & 26.10\textsl{\%} & 9.80\textsl{\%} & 0.3596 & 50.53\textsl{\%}   \\ 
LaD & 4.18\textsl{\%} & 18.53\textsl{\%} & 2.65\textsl{\%} & \textbf{0.1274} & \textbf{18.02\textsl{\%}}  \\
\hline
\end{tabular}  
}
\end{table}

\noindent \textbf{Performance on Toxic Test Set}

\noindent We further extract toxic prefixes from the KSQAC test dataset to create the Toxic Test Set, specifically designed to evaluate the model's detoxification capability. Meanwhile, we report the generation metrics for reference in Table \ref{tabel:KSQAC_toxic}. Although the model may achieve higher scores on generation metrics, it also introduces a greater number of toxic queries, significantly undermining the user experience in real-world applications. As shown in Table \ref{tabel:KSQAC_toxic}, \textbf{LaD} achieves the best performance on detoxification metrics. It indicates that \textbf{LaD} can significantly reduce the toxicity of the generated data. Meanwhile, rejecting toxic content during generation results in a 4.18\textsl{\%} value on R\textsl{@}4. This outcome is expected and enhances the user experience.

\subsection{Ablation Study}
\noindent To prove the influence of the essential components of LaD, we conduct extensive experiments on KSQAC Test datasets.

\noindent \textbf{The Effectiveness of Long-Short Interests 
 Hierarchical Capturing}

\begin{table}[t] 
\centering
\caption{Ablation study of Long-Short Interests 
 Hierarchical Capturing, where ``S'' denotes the length of short-term interests, ``L''  indicates the length of  long-term interests and ``{token nums}'' represents the average length of input tokens.}
\label{tabel:ablation_sl} 
\resizebox{0.85\columnwidth}{!}{
\begin{tabular}{l|ccc|ccc} 
  
\hline 
  & S & L & token nums &  R\textsl{@}4 $\uparrow$
 & BLEU $\uparrow$
 & MRR $\uparrow$
 \\
\hline  
\hline 
 SL-00 & 0 & 0 & 10 &  26.11\textsl{\%} & 31.95\textsl{\%} & 20.27\textsl{\%} \\
\hline  
\hline
{SL-10} & 1 & 0 & 20 & 29.22\textsl{\%} & 33.78\textsl{\%} & 22.53\textsl{\%}  \\
{SL-30} & 3 & 0 & 40 & 30.90\textsl{\%} & 34.40\textsl{\%} & 23.81\textsl{\%} \\
{SL-50} & 5 & 0 & 60 & \textbf{31.27\textsl{\%}} & \textbf{34.88\textsl{\%}} & \textbf{24.29\textsl{\%}} \\
\hline  
\hline 
{SL-01} & 0 & 1 & 11 & 28.35\textsl{\%} & 32.87\textsl{\%} & 21.80\textsl{\%} \\
{SL-03} & 0 & 3 & 13 & 29.77\textsl{\%} & 33.67\textsl{\%} & 23.04\textsl{\%} \\
{SL-05} & 0 & 5 & 15 & 30.03\textsl{\%} & 33.61\textsl{\%} & 23.37\textsl{\%} \\
{SL-020} & 0 & 20 & 30 & \textbf{30.58\textsl{\%}} & \textbf{34.23\textsl{\%}} & \textbf{24.07\textsl{\%}} \\
{SL-030} & 0 & 30 & 40 & 30.35\textsl{\%} & 34.20\textsl{\%} & 24.02\textsl{\%} \\
\hline  
\hline  
{SL-37} & 3 & 7 & 47 & \textbf{31.70\textsl{\%}} & \textbf{35.15\textsl{\%}} & \textbf{24.78\textsl{\%}} \\
\hline  
\end{tabular}  
}
\end{table} 

\noindent We conducted an experiment to investigate the impact of different quantities of short-term and long-term interests on the performance of generation. Incorporating more interests leads to longer text inputs and higher online inference time. The maximum token number of the prefix is 10, and each short-term interest is encoded into a sequence of no more than 10 tokens. The representation of long-term interests is pre-calculated and cached when applied online, so each long-term interest only increases the sequence length by 1. Thus, adding 1 short-term interest would result in the same increase in inference time as increasing 10 long-term interests.
The results are shown in Table \ref{tabel:ablation_sl}, the key observations and analysis are as follows: 

(1) \textbf{SL-00} lacks personalized information, resulting in the lowest performance in the metrics. This underscores the significance of personalized information in QAC tasks. 

(2) \textbf{SL-10}, \textbf{SL-30}, and \textbf{SL-50} are models that incorporate short-term interests. As the number of short-term interests increases (from 1 to 5), the R\textsl{@}4, BLEU, and MRR metrics improve by 2.05\textsl{\%}, 1.1\textsl{\%}, and 1.76\textsl{\%}, respectively. It suggests that short-term interests are effective in understanding the user's search intentions. 

(3) The models \textbf{SL-01}, \textbf{SL-03}, \textbf{SL-05}, \textbf{SL-020}, and \textbf{SL-030} are designed to develop long-term interests. As the duration of the long-term component extends, all metrics initially exhibit an increase, followed by a subsequent decrease. Our analysis suggests that more historical behavioral information can lead to better results, but too much information can also introduce noise.

(4) \textbf{SL-37}, meticulously crafted by us, delivers optimal results. Moreover, it more effectively demonstrates the practical application of generative models in real-world online QAC scenarios.
Compared to models with only short-term interests or only long-term interests, \textbf{SL-37} delivers superior results under similar text input length. Therefore, we use 3 short-term interests and 7 long-term interests to achieve a balance between performance and online latency.



\noindent \textbf{The Effectiveness of Adaptive Detoxification}

\begin{table}[t] 
\centering
\caption{Ablation study of Adaptive Detoxification, where ``AD'' denotes adaptive detoxification.}
\label{tabel:ablation_ad} 
\resizebox{1\columnwidth}{!}{
\begin{tabular}{l|ccc|cc|c} 
  
\hline 
  & R\textsl{@}4 $\uparrow$
 & BLEU $\uparrow$
 & MRR $\uparrow$
 &  UAmaxT $\downarrow$
 & UProb $\downarrow$
 & AvgRN \\
\hline  
\hline 
LaD w/o AD & \textbf{31.70\textsl{\%}} & \textbf{35.15\textsl{\%}} & \textbf{24.78\textsl{\%}} & 0.1400 &  19.31\textsl{\%}  &  0.0 \\ 
\hline  
\hline
LDPO & 30.19\textsl{\%} & 33.88\textsl{\%} & 23.77\textsl{\%} & 0.0837 & 11.01\textsl{\%} & 0.0 \\
LDPO w Offline Reject & 29.04\textsl{\%} & 32.96\textsl{\%}  & 22.99\textsl{\%} &  0.0881 & 11.47\textsl{\%} & 0.2976 \\
LDPO w Online Reject & 30.62\textsl{\%} & 33.76\textsl{\%}  & 24.07\textsl{\%} & 0.0753 & 9.51\textsl{\%} & 0.0125 \\
\hline  
\hline
LaD & 29.42\textsl{\%} & 31.77\textsl{\%}  & 23.09\textsl{\%} & \textbf{0.0590} & \textbf{6.55\textsl{\%}} & 0.2001 \\
\hline
\end{tabular}  
}
\end{table} 

\noindent The scheme of adaptive detoxification with RPO is the foundation for deploying the generative personalized QAC in online applications. We consider the following ablation models to demonstrate the effectiveness of our adaptive detoxification method:  
(1) \textbf{LaD w/o AD}: the model without any detoxification ability, which is equal to SL-37 in Tabel \ref{tabel:ablation_sl}. 
(2) \textbf{LDPO}: the model initiates detoxification using DPO optimization. 
(3) \textbf{LDPO w Offline Reject}: LDPO with offline {\ttfamily [Reject]} injection (i.e. the inject operation is applied during the preprocessing of the data’s ground truth). 
(4) \textbf{LDPO w Online Reject}: LDPO with online {\ttfamily [Reject]} injection. 
(5) \textbf{LaD}: our complete model, which lacks a reference model compared to \textbf{LDPO w Online Reject}.

The results are shown in Table \ref{tabel:ablation_ad}, the key observations and analysis are as follows:  

(1) \textbf{LaD w/o AD} achieves the best performance on generation
metrics, yet yields the worst results on UAmaxT and UProb. It suggests that while the detoxification process might negatively impact certain generation metrics, it enhances the overall quality of the outputs. Detoxifying text is crucial for online QAC generation. 

(2) \textbf{LDPO} exceeds \textbf{LaD w/o AD} by 0.0563 and 8.3\textsl{\%} in UAmaxT and UProb, demonstrating the detoxification capabilities of DPO-related methods. However, \textbf{LDPO} tends to produce irrelevant context when given with a toxic prefix. 

(3) \textbf{LDPO w Offline Reject} improves the UAmaxT and UProb by 0.0128 and 1.96\textsl{\%}. It proves that {\ttfamily [Reject]} injection can further detoxification the text. Moreover, the injection of {\ttfamily [Reject]} enables adaptive detoxification tailored to various different inputs. 

(4) \textbf{LDPO w Online Reject} not only improves the  R\textsl{@}4, BLEU and MRR metric but also significantly reduces the  UAmaxT and UProb compared to \textbf{LDPO w Offline Reject}. It proves the superiority of the online strategy. 

(5) \textbf{LaD} achieves the best performance in UAmaxT and UProb, with only a slight decrease in the R\textsl{@}4, BLEU, and MRR metrics compared to \textbf{LaD w/o AD}. This demonstrates that \textbf{LaD} strikes a balance between generation quality and detoxification. Futhermore, \textbf{LaD} only rejects an average of 0.2 completions while achieves the best performance in detoxification.

\subsection{Case Study}

\begin{figure}
\centering
\includegraphics[width=7.4cm]{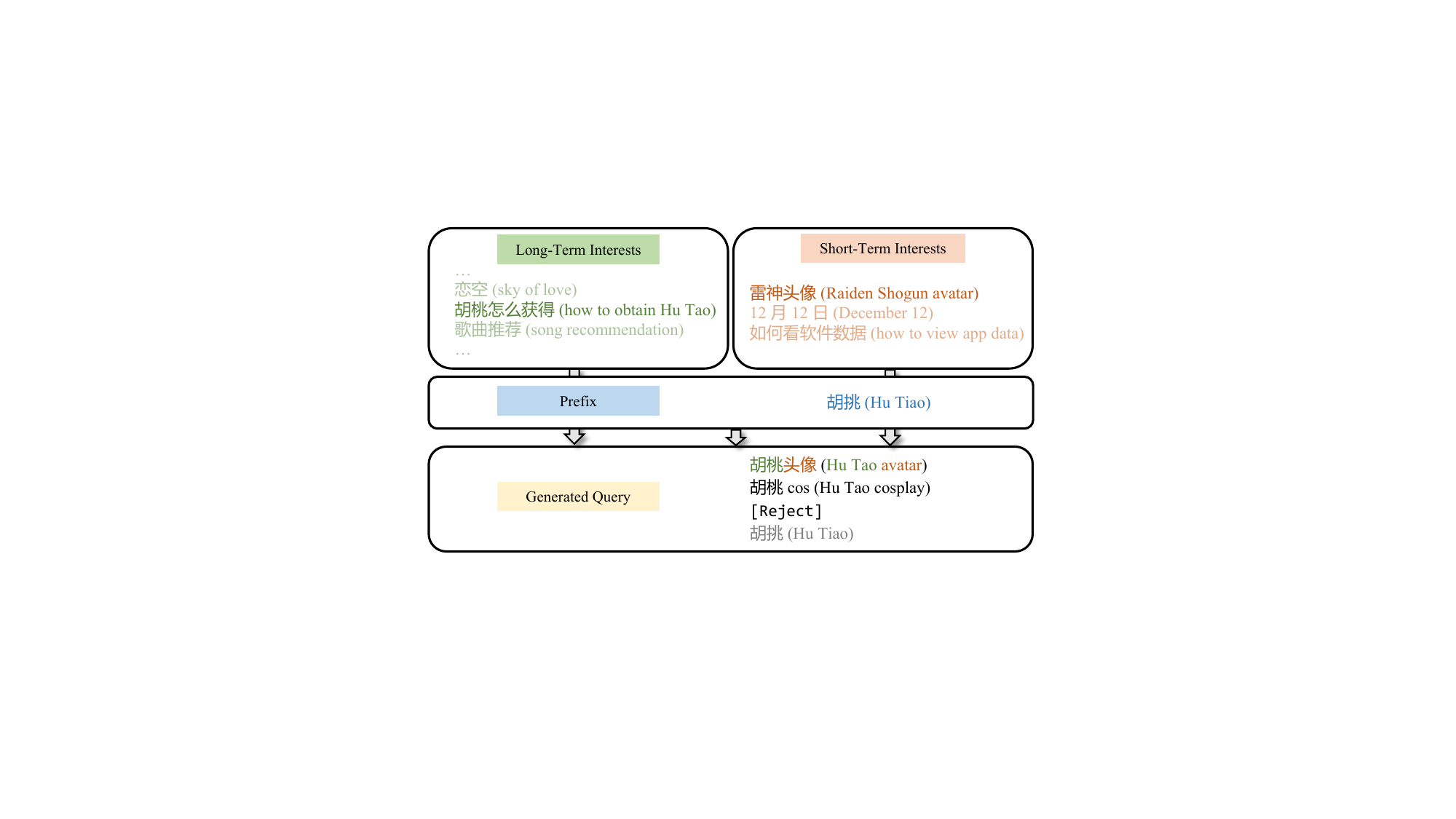}
\caption{Case study of our full model LaD. ``Hu Tao'' is a character in Genshin Impact.}
\label{fig:case}
\end{figure}

To make the analysis more explicit, we conduct case study. As shown in Figure \ref{fig:case}, the prefix has typos from ``Hu Tao'' to ``Hu Tiao''. Firstly, LaD can accurately capture information from long-term interests, such as ``how to obtain Hu Tao'', to determine that the user frequently searches for content related to the game Genshin Impact. Then, regarding instant interests, LaD can discern the intention related to the avatar from short-term interests, like ``Raiden Shogun avatar'', which is searched in proximity to the prefix. As the online result, the user clicked on the query, ``Hu Tao avatar'', we generated. Moreover, LaD can rank the special token {\ttfamily [Reject]} higher than the generated typos query ``Hu Tiao''. This generation will invisible to the user. This demonstrates that LaD can adaptively generate non-toxic text by leveraging both long-term and short-term interest information. Additional cases can be found in the Appendix.

\subsection{Online Testing}

As mentioned in Section \ref{sec:online_gene}, LaD is capable of delivering authentic real-time online inference. Here, we first integrate human feedback to evaluate the quality of the generations, and then deploy the model online to assess its effectiveness in real-world scenarios.

\begin{table}[t] 
\centering
\caption{The Human Study of Online A/B Testing. Due to the confidential nature of the KSQAC dataset, we report the the difference between the LaD and Baseline ($\triangle$  = LaD - Baseline).}
\label{tabel:ab_human} 
\resizebox{0.6\columnwidth}{!}{
\begin{tabular}{l|cc} 
  
\hline 
  & Baseline  & +LaD ($\triangle$) \\ 
\hline

Pornographic & - & -0.21\textsl{\%} \\
Misinformation & - & -0.04\textsl{\%}  \\
Typos & -  & -0.27\textsl{\%} \\
Irrelevant & -  & -0.10\textsl{\%}\\
Duplicate & - & -0.20\textsl{\%} \\
\hline 
Overall & - & -0.81\textsl{\%} \\

\hline
\end{tabular}  
}
\end{table} 

  



\noindent \textbf{Human Evaluation}

\noindent The model's generations are utilized as a new recall source and are integrated into the standard computation of subsequent pipelines. To assess the model's impact on the prior metrics of online performance, we conducted a manual evaluation on a total of 3,000 display data samples. We adopt Pornographic, Misinformation, Typos, Irrelevant and Duplicate as the metrics for human evaluation: 

\textbf{Pornographic}: It evaluates whether the generated query contains pornographic content. A higher value indicates a greater amount of pornographic content in the generated query.

\textbf{Misinformation}: It assesses the presence of negative or false information in the generated queries. A higher value signifies a larger number of queries containing such content.

\textbf{Typos}: It determines the presence of grammatical errors or typos in the generated queries. A higher value signifies a larger number of queries with these issues.

\textbf{Irrelevant}: It assesses whether the generated query is semantically related to the input prefix. A higher value indicates a greater number of generated queries that are unrelated to the prefix.

\textbf{Duplicate}: It assesses the presence of meaningless character repetitions in generated queries. A higher value suggests a larger number of queries containing such repetitions. 

The results are shown in Table \ref{tabel:ab_human}, LaD significantly reduces the bad case rate of completions by 0.81\textsl{\%}. It is attributed to the design of our detoxification strategy, which can adaptively generate detoxified text based on the prefix. Thanks to its exceptional adaptive detoxification capabilities, our model enables the safe deployment of real-time generation methods in industrial applications.

\noindent \textbf{Online A/B Testing}

\noindent We validate the proposed LaD with online A/B testing on Kuaishou search. We compare LaD with our latest production baseline. Both the experimental group and control group were randomly assigned 20\textsl{\%} of online users for a 15-day A/B test. As shown in Tabel \ref{tabel:ab_res}, the online evaluation metrics are divided into three categories. 

(1) \textbf{The} first part measures the contribution to the online QAC task itself. It includes three key metrics: CTR, PV and Second-day Retention Rate. CTR assesses whether the generated content meets user needs and drives click behavior. PV examines whether the model stimulates greater user search intent, resulting in an increased number of searches. Second-day Retention Rate assesses whether the generated content encourages users to return and use the QAC functionality again the next day. The results are shown in the first block of Table \ref{tabel:ab_res}. Our comprehensive model, LaD (Gen + AD + LS), contributes to a 4.08\textsl{\%} increase in CTR, a 4.398\textsl{\%} increase in PV, and a 0.646\textsl{\%} increase in the Second-day Retention Rate. This represents the largest CTR increase achieved in a single experiment over the past two years. 
(2) \textbf{The} second part evaluates user interaction with the search results page after clicking on the recommended query, focusing on Play Duration and Long-play Count. Play Duration measures the total time users spend watching videos, while Long-play Count tracks the number of times users watch a video for an extended period. As shown in the second block of Table \ref{tabel:ab_res}, LaD has stimulated extensive user engagement on the search results page. 
(3) \textbf{The} third part evaluates the model's impact on the overall search system. The Search Penetration Rate metric quantifies the proportion of users who utilize the search feature across the entire app. As shown in the third block of Table \ref{tabel:ab_res}, LaD improves the Search Penetration Rate by 0.438\textsl{\%}, indicating that it significantly expands the user base for search and plays a crucial role in cultivating users' search habits. 

\begin{table}[t] 
\centering
\caption{The improvements of LaD in online A/B test compared to the production baseline, where ``Gen'' denotes generative model, ``AD'' indicates adaptive detoxification and ``LS'' means long and short-term interests capturing. Text in grey indicates that the value is not significant.}
\label{tabel:ab_res} 
\resizebox{0.91\columnwidth}{!}{
\begin{tabular}{l|c|cc} 
  
\hline 
  & Nearline & Gen + AD  & Gen + AD + LS  \\
\hline  
\hline 
Click-through Rate (CTR) & +0.841\textsl{\%} & +2.48\textsl{\%} & +4.08\textsl{\%} \\
Search Num (PV) & +0.869\textsl{\%} & +2.745\textsl{\%} & +4.398\textsl{\%}   \\
Second-day Retention Rate & \color{gray}{+0.069\textsl{\%}} & +0.323\textsl{\%}  &  +0.646\textsl{\%}  \\
\hline  
\hline  
Play Duration & +0.905\textsl{\%} & +1.655\textsl{\%} & +3.655\textsl{\%} \\
Long-play Count & +0.634\textsl{\%} & +2.072\textsl{\%} & +4.554\textsl{\%} \\
\hline  
\hline
Search Penetration Rate & \color{gray}{+0.088\textsl{\%}} & +0.254\textsl{\%} & +0.438\textsl{\%} \\ 
\hline
\end{tabular}  
}
\end{table} 

LaD (Gen + AD + LS) outperforms Gen + AD across all metrics, highlighting the superiority of our designed hierarchical personalized capture method. 
Furthermore, the Nearline model (generate completions based on previous prefixes and build a daily query database \cite{yin2020learning}), with three times the number of parameters compared to our online model, achieved only minimal improvement. This is because the daily database update format fails to capture users' real-time dynamic needs, and the coverage of prefixes is relatively low. It highlights the superiority of online generation. 
\textbf{LaD} surpassed our highly optimized production baseline, delivering the largest single-experiment metric improvement in nearly two years. It has since been deployed on Kuaishou search, powering the primary traffic for hundreds of millions of active users.

\section{Conclusion}

In this paper, we introduce a new personalized QAC model, LaD, which integrates generative QAC with hierarchical capturing of long and short-term interests, along with adaptive detoxification. LaD delivers authentic online real-time inference. It surpasses a strong production baseline and has been deployed on Kuaishou search, handling primary traffic for hundreds of millions of active users. We consider applying our model to multilingual scenarios, exploring QAC for cold-start users as part of our future work. Furthermore, LaD can be developed with stream-based training, potentially replacing the recall, pre-ranking, and ranking stages by utilizing a single generative model for online serving.


\bibliographystyle{ACM-Reference-Format}
\bibliography{sample-base}

\appendix

\section{Appendix}

\subsection{Results on Public Dataset}

\noindent We also inplemented LaD on the AOL dataset \cite{pass2006picture}, which is the largest publicly available search log and is widely used in academic research on query auto completion. The AOL dataset contains three months of user search behavior logs, from March 1, 2006, to May 31, 2006. It includes information such as user IDs, queries, and time-stamp details. Keeping consistent with existing studies \cite{maheswaran2024dac}, we split the AOL dataset into three parts - trainset, validset, and testset - based on the order of time. The earlier data was used for model training and the later data was used as the testset to prevent data leakage. \textbf{The training set was comprised of two parts}: a larger dataset with 3 million queries used for training the base personalized generative model, and a smaller dataset with 90 thousands samples used for the second stage training aimed at reducing the risk of generating toxic queries. These two parts of the training dataset do not overlap. 

We use the Detoxify\footnote{\url{https://github.com/unitaryai/detoxify}} tool, which is a publicly available toxic comment classification model, to calculate the toxicity score. A score greater than 0.5 indicates that a query is considered toxic. Based on this, the testset is split into two parts, $T_{test}$ indicating toxic and $NT_{test}$ indicating non-toxic, to evaluate the toxicity of the model's generated outputs. We use beam search algorithm with beam size 10 to generate 10 results for each session. In order to make a fair comparison, all baselines are inplemented on this dataset, and the results are presented in the Table \ref{tabel:aol_toxic} and Table \ref{tabel:aol_nontoxic}. The key observations and analysis are as follows: 
(1) \textbf{The Effectiveness of More Data.} Following the convention, the model \textbf{LaD w/o AD} is trained solely on the first stage data, because the second stage training set is specifically designed to minimize the risk of generating toxic queries. For the Non-Toxic Test Set (as shown in Table \ref{tabel:aol_nontoxic}), \textbf{LaD} outperforms \textbf{LaD w/o AD} on generation metrics indicating that additional data enhances performance. For the Toxic Test Set (as shown in Table \ref{tabel:aol_toxic}), \textbf{LaD} outperforms \textbf{LaD w/o AD} in terms of UAmaxT and UProb, while it results in a decrease in the generation metrics. This is because \textbf{LaD} uses an adaptive detoxification strategy, generating completions more cautiously when faced with toxic prefixes. By refusing to produce toxic text, it consequently reduces the generation metrics. 
(2) \textbf{Superior Detoxification Performance of LaD.} \textbf{LaD} achieves the best performance on detoxification metrics, excelling on both toxic and non-toxic test sets. This highlights that our model possesses superior detoxification capabilities. 
(3) \textbf{Superior Generative Performance.} For non-toxic prefixes, shown in Table \ref{tabel:aol_nontoxic}, LaD achieves the best performance on all metrics. While adaptively detoxifying for toxic prefixes, the model also exhibits excellent generative capabilities for non-toxic prefixes. This may be attributed to our hierarchical personalized representation design.

\begin{table}[t] 
\centering
\caption{Performance comparison for Toxic Test Set ($T_{test}$) of AOL dataset, where ``AD'' denotes adaptive detoxification.}
\label{tabel:aol_toxic} 
\resizebox{0.95\columnwidth}{!}{
\begin{tabular}{l|ccc|cc} 
  
\hline 
  & R\textsl{@}4 $\uparrow$
 & BLEU $\uparrow$
 & MRR $\uparrow$
 &  UAmaxT $\downarrow$ & UProb $\downarrow$ \\
\hline  
\hline 
MPC \cite{bar2011context} & 11.05\textsl{\%} & 12.18\textsl{\%} & 8.43\textsl{\%} & 0.2496 & 28.90\textsl{\%}  \\
Gen \cite{mustar2020using} & 28.72\textsl{\%} & 36.53\textsl{\%} & 24.23\textsl{\%} & 0.1623 & 17.31\textsl{\%}  \\
M$^{2}$A \cite{yin2020learning} & 32.20\textsl{\%} & 37.89\textsl{\%} & 27.47\textsl{\%} & 0.1831 & 19.54\textsl{\%}  \\
SIN \cite{bao2024search}  & 32.88\textsl{\%} & 38.57\textsl{\%} & 27.76\textsl{\%} & 0.1855 & 19.81\textsl{\%}  \\
Trie-NLG \cite{maurya2023trie}  & 31.01\textsl{\%} & 37.72\textsl{\%} & 26.61\textsl{\%} & 0.1832 & 19.88\textsl{\%}  \\

\hline  
\hline
DAPT \cite{gururangan2020don}  & 37.89\textsl{\%} & 46.23\textsl{\%} & 33.43\textsl{\%} & 0.1735 & 18.18\textsl{\%}  \\
Quark \cite{lu2022quark}  & 29.06\textsl{\%} & 35.11\textsl{\%} & 22.92\textsl{\%} & 0.1662 & 17.71\textsl{\%}  \\
PPO \cite{ziegler2019fine} & 36.02\textsl{\%} & 43.00\textsl{\%} & 30.82\textsl{\%} & 0.1739 & 18.61\textsl{\%}  \\
DAC \cite{maheswaran2024dac}  & 27.61\textsl{\%} & 36.64\textsl{\%} & 22.54\textsl{\%} & 0.1544 & 16.60\textsl{\%}  \\
\hline  
\hline
LaD w/o AD & 33.90\textsl{\%} & 43.27\textsl{\%} & 29.09\textsl{\%} & 0.1706 & 18.24\textsl{\%}   \\ 
LaD & 16.99\textsl{\%} & 29.82\textsl{\%} & 14.19\textsl{\%} & \textbf{0.0923} & \textbf{6.97\textsl{\%}}  \\
\hline
\end{tabular}  
}
\end{table}

\begin{table}[t] 
\centering
\caption{Performance comparison for Non-Toxic Test Set ($NT_{test}$) of AOL dataset, where ``AD'' denotes adaptive detoxification.}
\label{tabel:aol_nontoxic} 
\resizebox{1\columnwidth}{!}{
\begin{tabular}{l|ccc|cc} 
  
\hline 
  & R\textsl{@}4 $\uparrow$
 & BLEU $\uparrow$
 & MRR $\uparrow$
 &  UAmaxT $\downarrow$ & UProb $\downarrow$ \\
\hline  
\hline 
MPC \cite{bar2011context} & 21.45\textsl{\%} & 24.15\textsl{\%} & 18.43\textsl{\%} & 0.0308 & 2.10\textsl{\%}  \\
Gen \cite{mustar2020using} & 33.60\textsl{\%} & 43.56\textsl{\%} & 28.95\textsl{\%} & 0.0271 & 1.54\textsl{\%}  \\
M$^{2}$A \cite{yin2020learning} & 33.59\textsl{\%} & 44.10\textsl{\%} & 28.92\textsl{\%} & 0.0266 & 1.45\textsl{\%}  \\
SIN \cite{bao2024search}  & 34.06\textsl{\%} & 44.52\textsl{\%} & 29.32\textsl{\%} & 0.0263 & 1.40\textsl{\%}  \\
Trie-NLG \cite{maurya2023trie}  & 31.40\textsl{\%} & 41.82\textsl{\%} & 27.47\textsl{\%} & 0.0310 & 1.93\textsl{\%}  \\

\hline  
\hline
DAPT \cite{gururangan2020don}  & 41.73\textsl{\%} & 51.35\textsl{\%} & 36.73\textsl{\%} & 0.0206 & 0.69\textsl{\%}  \\
Quark \cite{lu2022quark}  & 31.12\textsl{\%} & 39.26\textsl{\%} & 24.51\textsl{\%} & 0.0173 & 0.68\textsl{\%}  \\
PPO \cite{ziegler2019fine} & 38.38\textsl{\%} & 48.21\textsl{\%} & 32.96\textsl{\%} & 0.0221 & 0.92\textsl{\%}  \\
DAC \cite{maheswaran2024dac}  & 30.42\textsl{\%} & 41.77\textsl{\%} & 24.74\textsl{\%} & 0.0182 & 0.69\textsl{\%}  \\
\hline  
\hline
LaD w/o AD & 36.64\textsl{\%} & 48.42\textsl{\%} & 32.03\textsl{\%} & 0.0238 & 1.14\textsl{\%}   \\ 
LaD & 41.73\textsl{\%} & 51.42\textsl{\%} & 36.61\textsl{\%} & \textbf{0.0167} & \textbf{0.23\textsl{\%}}  \\
\hline
\end{tabular}  
}
\end{table}

\subsection{Case Study}

\textbf{The Effectiveness of LaD}

\begin{figure}[t]
\centering
\includegraphics[width=7.5cm]{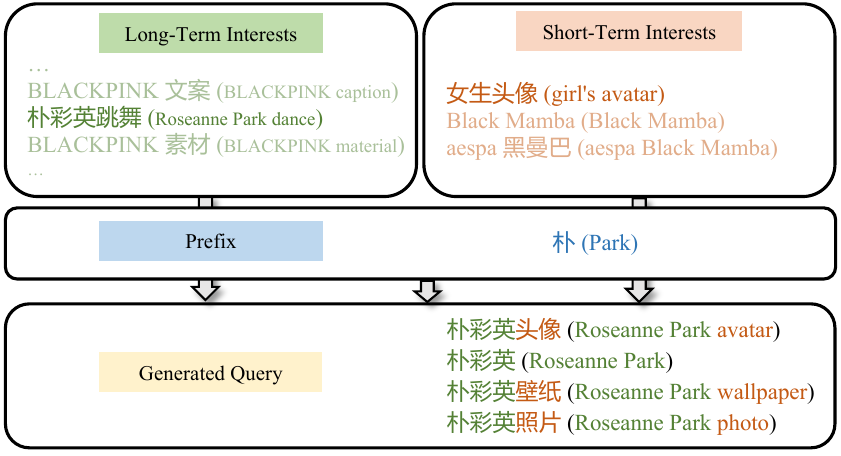}
\caption{Case study of our full model LaD. ``Roseanne Park'' is a Korean-New Zealand singer and dancer, and is a member of the South Korean girl group BLACKPINK.}
\label{fig:exp_app_case2}
\end{figure}

\noindent  As shown in Figure \ref{fig:exp_app_case2}, the prefix entered by the user is a surname ``Park''. Long-term interests indicate that the user is a fan of {\it Roseanne Park }, who is a member of the South Korean girl group BLACKPINK. The user's recent search activity indicates a short-term interest in information related to avatars. Thus, LaD produces impressive outputs, such as Roseanne Park avatars, wallpapers, and photos. Note that the special token {\ttfamily [Reject]} is not generated by LaD, as both the prefix and completions are already non-toxic.

\noindent \textbf{The Effectiveness of Adaptive Detoxification}

\begin{figure}[t]
\centering
\includegraphics[width=7cm]{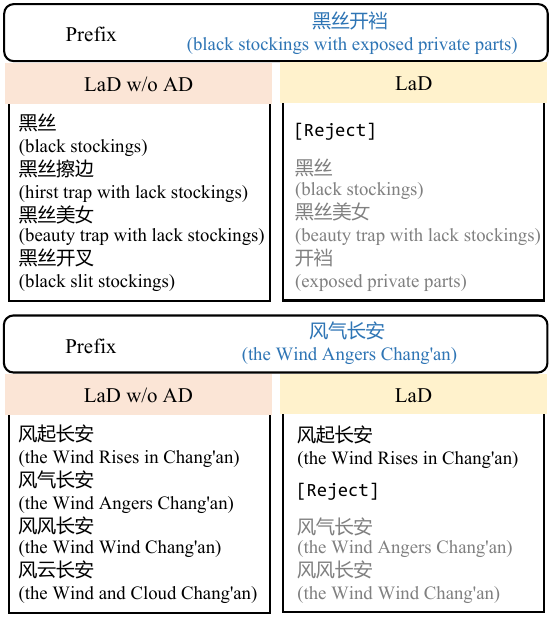}
\caption{Case study of adaptive detoxification, we omit the display of long and short interests contents. ``AD'' indicates adaptive detoxification.}
\label{fig:exp_app_case1}
\end{figure}

\noindent As shown in Figure \ref{fig:exp_app_case1}, we show different types of the input prefix: (1) The first type of prefix contains pornographic sensitive information: ``black stockings with exposed private parts''. \textbf{LaD} generates {\ttfamily [Reject]} at the top position, ensuring that the pornographic text remains invisible to the user. In contrast, \textbf{LaD w/o AD} generates responses with pornographic content, such as ``hirst trap with lack stockings'', which is not suitable for display. (2) The second type of prefix contains typographical errors: ``the Wind {\it Angers} Chang'an'' (where ``the Wind {\it Rises} Chang'an'' is a Chinese novel). \textbf{LaD} generates {\ttfamily [Reject]} at the second position. The generation that ranks higher than {\ttfamily [Reject]} is perfect. The generation that ranks lower than {\ttfamily [Reject]} contains some typos, such as the repeated phrase ``Wind Wind''. With the {\ttfamily [Reject]}, generations ranked lower than {\ttfamily [Reject]} will be discarded. However, \textbf{LaD w/o AD} continues to produce generations with typos, such as ``Wind and Cloud Chang'an''.

\textbf{LaD} can adaptively generate non-toxic content based on different types of prefixes. This capability is crucial for platform safety and user experience. Moreover, the toxicity of completions returned to users is a critical metric in practice. Adaptive detoxification allows generative models to be more effectively deployed in online environments. \textbf{LaD} is an end-to-end detoxification approach, where the detoxification capability is learned during training, without introducing any additional processing time.







 \end{document}